%% file: main.tex
\documentclass[sigconf]{acmart}

\usepackage{lipsum}
\usepackage{subcaption}
\usepackage{graphicx}
\usepackage{hyperref}
\usepackage{threeparttablex}
\usepackage{multirow}
\usepackage{bm}
 \usepackage{balance}

\newcommand{\boldres}[1]{{\textbf{\textcolor{red}{#1}}}}
\newcommand{\secondres}[1]{{{#1}}}
\newcommand\Tstrut{\rule{0pt}{2.6ex}}         % = `top' strut

\copyrightyear{2023}
\acmYear{2023}
\setcopyright{rightsretained}
\acmConference[CIKM '23]{Proceedings of the 32nd ACM International Conference on Information and Knowledge Management}{October 21--25, 2023}{Birmingham, United Kingdom}
\acmBooktitle{Proceedings of the 32nd ACM International Conference on Information and Knowledge Management (CIKM '23), October 21--25, 2023, Birmingham, United Kingdom}\acmDOI{10.1145/3583780.3615159}
\acmISBN{979-8-4007-0124-5/23/10}

\makeatletter
\gdef\@copyrightpermission{
  \begin{minipage}{0.3\columnwidth}
   \href{https://creativecommons.org/licenses/by/4.0/}{\includegraphics[width=0.90\textwidth]{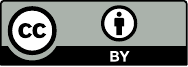}}
  \end{minipage}\hfill
  \begin{minipage}{0.7\columnwidth}
   \href{https://creativecommons.org/licenses/by/4.0/}{This work is licensed under a Creative Commons Attribution International 4.0 License.}
  \end{minipage}
  \vspace{5pt}
}
\makeatother

\begin{document}

\title{Unlocking the Potential of Deep Learning in Peak-Hour Series Forecasting}
% Bridge the Performance Gap in Peak-hour Series Forecasting: The Seq2Peak Framework

\author{Zhenwei Zhang}
\affiliation{%
  \department{Department of Electronic Engineering}
  \institution{Tsinghua University}
  \city{Beijing}
  \country{China}
}
\email{zzw20@mails.tsinghua.edu.cn}

\author{Xin Wang}
\affiliation{%
  \department{Department of Electronic Engineering}
  \institution{Tsinghua University}
  \city{Beijing}
  \country{China}
}
\email{wangxin20@mails.tsinghua.edu.cn}

\author{Jingyuan Xie}
\affiliation{%
  \department{Department of Electronic Engineering}
  \institution{Tsinghua University}
  \city{Beijing}
  \country{China}
}
\email{xiejy20@mails.tsinghua.edu.cn}

\author{Heling Zhang}
\affiliation{%
  \department{Department of Electronic Engineering}
  \institution{Tsinghua University}
  \city{Beijing}
  \country{China}
}
\email{zhanghel20@mails.tsinghua.edu.cn}

\author{Yuantao Gu}
\affiliation{%
  \department{Department of Electronic Engineering}
  \institution{Tsinghua University}
  \city{Beijing}
  \country{China}
}
\email{gyt@tsinghua.edu.cn}


\begin{abstract}
Unlocking the potential of deep learning in Peak-Hour Series Forecasting (PHSF) remains a critical yet underexplored task in various domains. While state-of-the-art deep learning models excel in regular Time Series Forecasting (TSF), they struggle to achieve comparable results in PHSF. This can be attributed to the challenges posed by the high degree of non-stationarity in peak-hour series, which makes direct forecasting more difficult than standard TSF. Additionally, manually extracting the maximum value from regular forecasting results leads to suboptimal performance due to models minimizing the mean deficit.  To address these issues, this paper presents Seq2Peak, a novel framework designed specifically for PHSF tasks, bridging the performance gap observed in TSF models. Seq2Peak offers two key components: the CyclicNorm pipeline to mitigate the non-stationarity issue and a simple yet effective trainable-parameter-free peak-hour decoder with a hybrid loss function that utilizes both the original series and peak-hour series as supervised signals. Extensive experimentation on publicly available time series datasets demonstrates the effectiveness of the proposed framework, yielding a remarkable average relative improvement of 37.7\% across four real-world datasets for both transformer- and non-transformer-based TSF models.
\end{abstract}



\begin{teaserfigure}
    \begin{subfigure}[b]{0.705\textwidth}   
        \centering 
        \includegraphics[width=\textwidth]{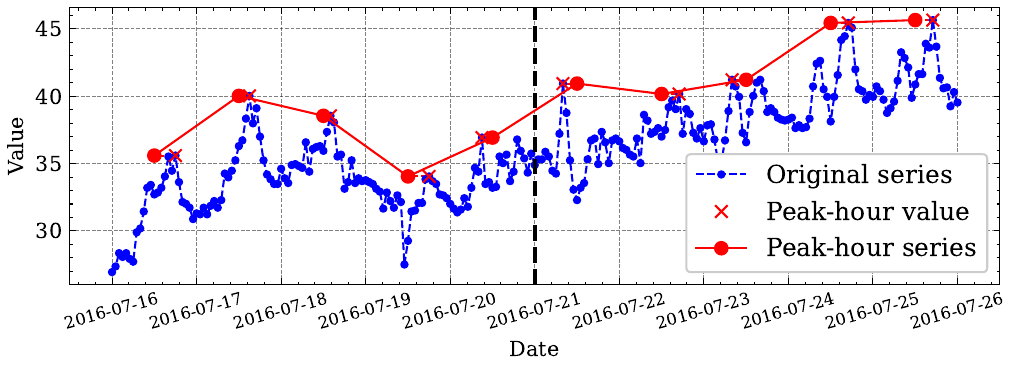}
        \caption[]{{\small Original time series and peak-hour series}}    
        \label{fig:peakhour}
    \end{subfigure}
    \hfill
    \begin{subfigure}[b]{0.29\textwidth}   
        \centering 
        \includegraphics[width=\textwidth]{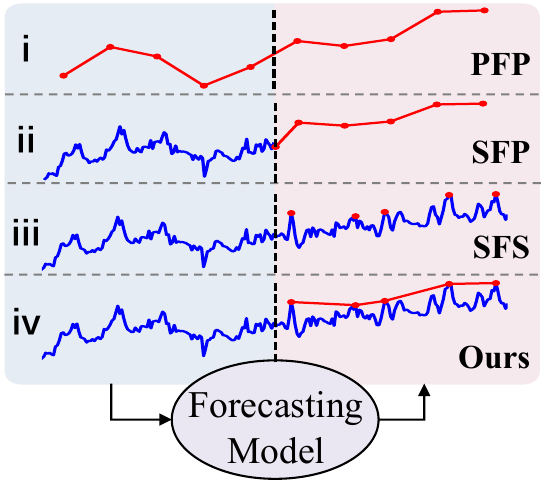}
        \caption[]{{\small PHSF paradigms}}    
        \label{fig:4frameworks}
    \end{subfigure}
  \vspace{-5pt}
  \caption{The peak-hour series forecasting task and its paradigms}
  \label{fig:teaser}
\end{teaserfigure}

\maketitle

% \begin{figure*}[t]
%     \begin{subfigure}[b]{0.705\textwidth}   
%         \centering 
%         \includegraphics[width=\textwidth]{imgs/peak_hour.pdf}
%         \caption[]{{\small Original time series and peak-hour series}}    
%         \label{fig:peakhour}
%     \end{subfigure}
%     \hfill
%     \begin{subfigure}[b]{0.29\textwidth}   
%         \centering 
%         \includegraphics[width=\textwidth]{imgs/4schemes.pdf}
%         \caption[]{{\small PHSF paradigms}}    
%         \label{fig:4frameworks}
%     \end{subfigure}
%   % \includegraphics[width=\textwidth]{imgs/peak_hour.pdf}
%   \vspace{-5pt}
%   \caption{The peak-hour series forecasting task and its paradigms}
%   % \Description{Enjoying the baseball game from the third-base
%   % seats. Ichiro Suzuki preparing to bat.}
% \end{figure*}

\section{Introduction}
Variations in peak-hour values within daily time series cycles (see Figure \ref{fig:peakhour}) are crucial across domains. In telecommunications, engineers calibrate base station capacities with maximum traffic volumes, optimizing communication quality \cite{pimpinella2022forecasting, 9909347}. In energy, daily peak power consumption affects the provisioning of raw materials and power production \cite{lee2022national, kim2020peak}, with some utilities basing billing on peak demand. Likewise, understanding peak-hour traffic patterns in the transportation sector is crucial to effective urban planning \cite{8735916}. Hence, precise Peak-hour Series Forecasting (PHSF) is significant for diverse industries, underlining the necessity of focused research in this area.

 Despite the importance of PHSF, existing research is underdeveloped, often treating PHSF as a conventional sequence-to-sequence problem \cite{goia2010functional, garcia2011characterization, amin2008combined}, and ignoring the vital relationship between peak-hour and original series. These approaches often fall short in predictive scope and accuracy. Recent advancements in TSF techniques \cite{9909348}, such as LogTrans \cite{li2019enhancing}, Informer \cite{zhou2021informer}, SageFormer \cite{zhang2023sageformer}, and DLinear \cite{zeng2022transformers}, offer promise but have limitations when directly applied to PHSF (Figure \ref{fig:results}). They struggle with the specific information embedded in peak-hour values and weak correlations, emphasizing the need for an innovative method tailored for PHSF.

\begin{figure}[htb]
    \centering
    \includegraphics[width=0.9\linewidth]{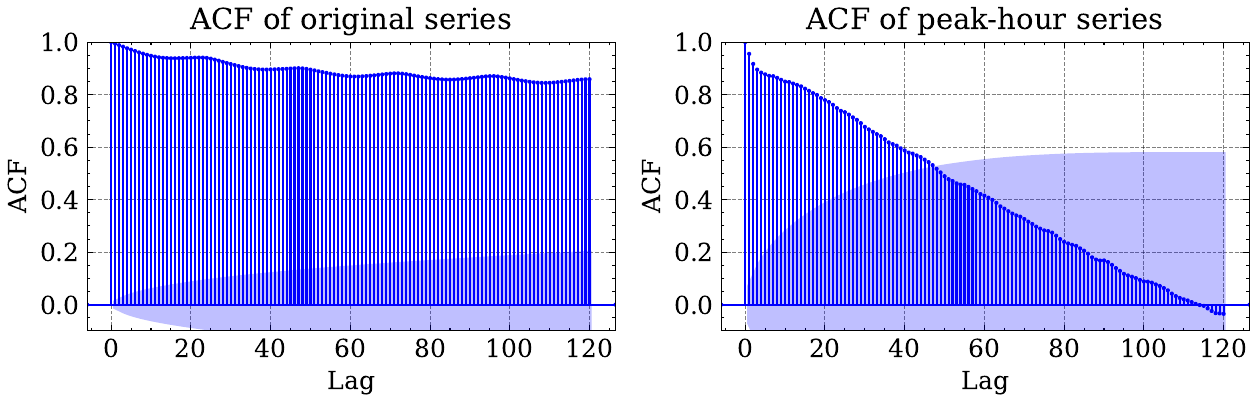}
    \caption{ACF for "OT" variable of ETTh1 \small{(5\% significance limits)}}
    \label{fig:acf}
\end{figure}

We identify three paradigms in applying TSF methodologies to PHSF tasks (Figure \ref{fig:4frameworks}). The first paradigm (PFP) solely relies on historical peak-hour series to forecast peak-hour series but often underperforms due to lower Autocorrelation Function (ACF) \cite{jenkins1957spectral} compared to the full series (Figure \ref{fig:acf}). Research has demonstrated a positive correlation between ACF and data predictability \cite{han2023capacity}. The second (SFP) employs full series to forecast but faces similar challenges. The third (SFS) incorporates full series and manually extracts maximum daily values, compromising the ability to predict extremes due to the loss function minimizing average error. For effective PHSF, it's necessary to: 1) Address wide forecast span and poor temporal dependencies; 2) Utilize the relationship between peak-hour and original series.

We introduce \textbf{Seq2Peak}, a framework bridging general time series and peak-hour series forecasting. It consists of two components: Cyclic Normalization, modeling inter-cycle relationships and extracting tailored statistical measures for PHSF; and Peak-hour Decoder, taking historical series, outputting original and peak-hour series with a hybrid loss function prioritizing the mapping relationship. Easily integrated into various models, Seq2Peak enhances forecasting without major computational complexity. Efficacy is proven by improved performance across models and real-world datasets, backed by ablation studies and parameter experiments, marking a new phase in the under-researched PHSF field.

The contributions of this paper are summarized as follows: 
1) We systematically introduce the task of peak-hour series forecasting, a vital, under-researched problem, highlighting challenges.
2) We propose Seq2Peak, a PHSF framework, easily integrated and generalized in most forecasting models
3) We validate our framework on four real-world datasets, showing 37.7\% average improvement over original TSF models.

\section{Related Works}
\paragraph{Peak-hour series forecasting:} Current forecasting methods primarily rely on traditional TSF techniques. Various studies have applied traditional TSF methods to predict peak time, including \cite{kwon2019weekly}, which uses LSTM\cite{shi2015convolutional} to forecast the peak load of the current week, and \cite{lee2022national}, which compares the performance of ARIMA\cite{zhang2003time}, SVR\cite{hastie2009elements}, LSTM on power load data. Similarly, \cite{pimpinella2022forecasting} utilizes ARIMA and LSTM, among other methods, to predict traffic patterns clustered for each base station during a week. \cite{yin2020forecast, mughees2021deep} introduces Bi-LSTM to forecast the day-ahead peak electricity. \cite{ibrahim2021deep} compares the performance of CNN and LSTM on peak-load data. \cite{goia2010functional} and \cite{nepal2020electricity} use linear regression and ARIMA, respectively, to model the relationship between the peak load of the next day and typical load patterns identified through clustering. However, with their sole reliance on historical peak time information and traditional time series prediction models, these methods often result in poor prediction performance. 
% Our work introduces powerful models such as the Transformer to peak time prediction tasks to respond to these shortcomings, promising an enhanced predictive capacity.

% \paragraph{Time series forecasting:} Recently, novel sequence prediction models, represented by the Transformer, have demonstrated capabilities surpassing those of LSTM, RNN, and other models. However, these models have not yet been widely applied to the peak time prediction domain. For example, Autoformer \cite{wu2021autoformer} incorporates time series decomposition and autocorrelation into the Transformer architecture, better extracting information from time series. Informer \cite{zhou2021informer} introduces sparsity into the attention mechanism, reducing the complexity of the Transformer, and making it more suitable for long-term prediction. Additionally, \cite{zeng2022transformers} proposes linear prediction models DLinear and NLinear, which have a much smaller scale than the Transformer while exhibiting comparable performance in prediction tasks. \cite{fan2022depts} presents a periodicity model based on residual learning, capable of effectively capturing various periodic characteristics in time series data. Our experiments show that directly applying these models to peak time prediction tasks yields unsatisfactory results. However, by combining these models with the peak time prediction framework proposed in this paper, their potential can be better harnessed in peak time prediction tasks.

\paragraph{Time series forecasting:} In recent years, innovative sequence processing models like the Transformer have outperformed traditional models like ARIMA and LSTM. Yet, their application in the domain of PHSF remains limited. Noteworthy developments in this area include SageFormer \cite{zhang2023sageformer}, which integrates Transformer and Graph Neural Network into the series-aware framework. Informer \cite{zhou2021informer} introduces sparsity into the attention mechanism to reduce the complexity of the Transformer. Additionally, \cite{zeng2022transformers} proposes DLinear, linear prediction models that perform comparably to the Transformer but are significantly smaller in scale. Unfortunately, our experiments indicate that directly applying these models to PHSF tasks yields unsatisfactory results. However, integrating these models with the proposed Seq2Peak framework can more effectively harness their potential, significantly advancing PHSF tasks.
% Moreover, \cite{fan2022depts} presents a periodicity model based on residual learning, which effectively captures various periodic characteristics in time series data. 

\section{Methodology}
\begin{figure}[t]
    \centering
    \includegraphics[width=0.85\linewidth]{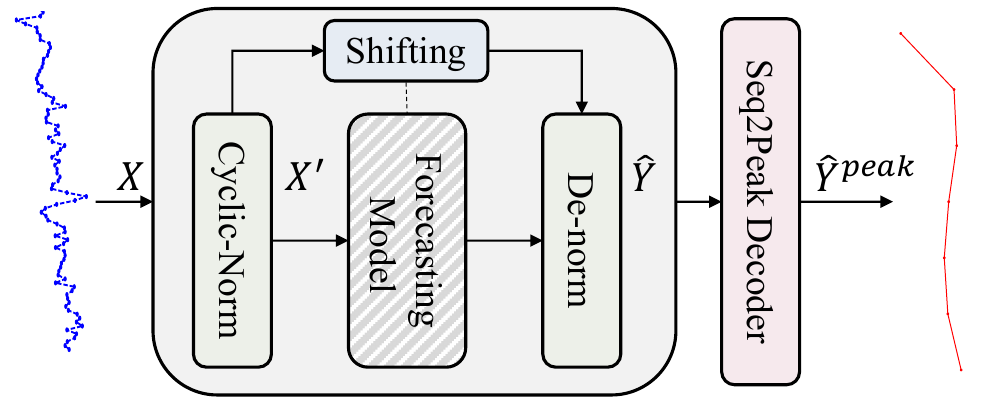}
    \caption{The proposed Seq2Peak framework}
    \label{fig:model}
\end{figure}

Given the limited research on PHSF tasks, we formally define them and introduce \textbf{Seq2Peak} (Figure \ref{fig:model}), a pioneering approach to PHSF. The framework includes two components: the \textbf{Cyclic Normalization} mechanism and the \textbf{Seq2Peak decoder}, working together to predict peak-hour series from historical data.

In this section, we formally address the definition of the Peak-hour Series Forecasting task. An illustrative diagram of this problem is provided in Figure \ref{fig:peakhour}. We denote a historical input data window of length $N$, represented as $X_t=\{\bm{x}_t,\bm{x}_{t+1},\cdots,\bm{x}_{t+N-1}\}$ ($t$ is omitted hereafter), where $\bm{x}_i\in\mathbb{R}^{c}$ and $c$ represents the number of channels. The objective is to forecast the peak-hour series $Y^{peak}$ derived from the original future series $Y=\{\bm{x}_{t+N},\cdots,\bm{x}_{t+N+M-1}\}$ with window size $M$.

The peak-hour value is the maximum value downsampled from an interval of $T=24$ hours consecutively for each time series channel.  Consequently, we obtain $Y^{peak} \in \mathbb{R}^{\frac{M}{T} \times c}$ by:
\begin{equation}
y^{\text{peak}}_{i,j}=max\{x_{(i-1)T+1,j},x_{(i-1)T+2,j},...,x_{iT,j}\},
\end{equation}
where $j$ denotes the channel index. Notably, peak-hour values for different channels may occur at different hours.

A PHSF model performs as a function to predict the future peak-hour series $Y^{\text{peak}}$ based on the historical full series $X$, 
\begin{equation}
    f(\cdot):\mathbb{R}^{N \times c} \rightarrow \mathbb{R}^{\frac{M}{T} \times c}.
\end{equation}
The predicted output is denoted as $\hat{Y}^{\text{peak}}=f(X)$.

\subsection{Cyclic Normalization}\label{sec:CyclicNorm}

The requirement for a longer time span and less self-correlation are the main characteristics of PHSF tasks. Our study addresses these concerns by introducing the Cyclic Normalization (CyclicNorm) pipeline, which aims to learn data correlation and generate features more appropriate for PHSF tasks. CyclicNorm achieves this by modeling the correlation of distribution across varying hours within a cyclical interval.

First, we perform Cyclic Normalization on the raw input time series to learn intrinsic correlations between different hours within the same cycle, separating non-stationary content. Given $X^{(i)}= \{\bm{x}_i,\bm{x}_{i+T}, \bm{x}_{i+2T},...\}$, where $1 \leq i \leq T$ and $i+2T < N$, we apply standard normalization $Norm(\cdot)$ to the $24$ sub-sequences individually, extracting means and dividing by standard deviations:
\begin{align}
\{X'^{(i)},\bm{\mu}_i,\bm{\sigma}_i\} = Norm(X^{(i)}).
\end{align}
Here, $\bm{M}=\{\bm{\mu}_1,\bm{\mu}_2,...,\bm{\mu}_T\}$ and $\bm{\Sigma}=\{\bm{\sigma}_1,\bm{\sigma}_2,...,\bm{\sigma}_T\}$ are the sets of means and standard deviations, respectively. The normalized $X'^{(i)}$ is the forecasting model's input for more stable data correlation.

The second stage of CyclicNorm further explores handling non-stationary input sequence statistics. Inspired by \cite{liu2022non, kim2021reversible}, we propose a non-stationary shifting module to model the distribution shifting in time series. 
\begin{align}
(\bm{M'}, \bm{\Sigma'}) = \mathcal{T}(\bm{M}, \bm{\Sigma})
\end{align}
Depending on the forecasting model and the dataset characteristics, the shifting function $\mathcal{T}$ can be instantiated as a trainable linear layer or simply a set of multiplication factors. 

The third stage of CyclicNorm involves denormalizing the results obtained from the forecasting model (i.e., the baseline models used in the experiments) using the aforementioned post-processed statistics $\bm{M'}$ and $\bm{\Sigma}'$. The output of CyclicNorm is treated as a standard TSF result, which also functions as an input for subsequent peak-hour inference.

\subsection{Seq2Peak Decoder}\label{sec:Seq2Peak}

The objective of optimization for standard TSF tasks focuses on minimizing the mean forecasting deficit, which contradicts the task of peak-hour series forecasting. However, PHSF models with a direct optimization strategy over peak-hour values have been proven to have poor generalizing ability to the test set. 

To overcome this dilemma, our Seq2Peak Decoder provides a simple yet highly effective optimization strategy: to optimize the loss function of the original time series and its corresponding peak-hour series simultaneously. To execute this strategy without introducing more trainable parameters, we attach a max-pooling layer of a stride and kernel size of 24 at the end of the previous standard forecasting result. 
The distinction between the max-pooling operation and manually processing the peak-hour series is that max-pooling allows back-propagation. Thus, we optimize the parameters of PHSF models via the following hybrid loss function $l_{hy}$.
% \begin{equation}
% \hat{Y}^{\text{peak}}_{i,j}=\max\{\hat{Y}_{(i-1)T+1,j},\hat{Y}_{(i-1)T+2,j},...,\hat{Y}_{iT,j}\}.
% \end{equation}
%To optimize the model's peak-time forecasting performance, the Seq2Peak Decoder takes a hybrid loss as its loss function:
\begin{equation}
l_{\text{hy}}=\alpha l_{\text{seq}}+(1-\alpha)l_{\text{peak}},
\label{loss func}
\end{equation}
where $l_{seq}$ is the MSE loss between the ground truth original time series and the output of the penultimate layer, $l_{peak}$ is the MSE loss between the ground truth of peak-hour series and the final output of the Seq2Peak decoder. $\alpha$ is a weighting factor that varies between 0 and 1. By employing this decoder and corresponding loss function, forecasting models can achieve stronger generalization abilities, fully capturing the original series' information while emphasizing forecasting performance for peak hour series.

\input{tables/compare.tex}

\section{Experiments}

\subsection{Experimental Setup}
\paragraph{Datasets} We evaluate our methods on four large-scale real-world time series datasets: ETTh1/2\footnote{https://github.com/zhouhaoyi/ETDataset}, Electricity\footnote{https://archive.ics.uci.edu/ml/ datasets/ElectricityLoadDiagrams20112014}, and Traffic\footnote{http://pems.dot.ca.gov}. These datasets have an hourly granularity and belong to the domains of energy and traffic. They exhibit daily periodicity, making the PHSF task meaningful in this context.

\paragraph{Baselines} We apply Seq2peak to transformer-based and non transformer-based TSF models (Transformer \cite{vaswani2017attention}, Informer \cite {zhou2021informer}, Autoformer \cite{wu2021autoformer}, and DLinear \cite{zeng2022transformers}) to investigate performance enhancement. For baselines, we perform the SFS paradigm. We examine mean square error (MSE) and mean absolute error (MAE) as metrics.

\begin{figure}[thbp]    
    \centering
    \includegraphics[width=0.9\columnwidth]{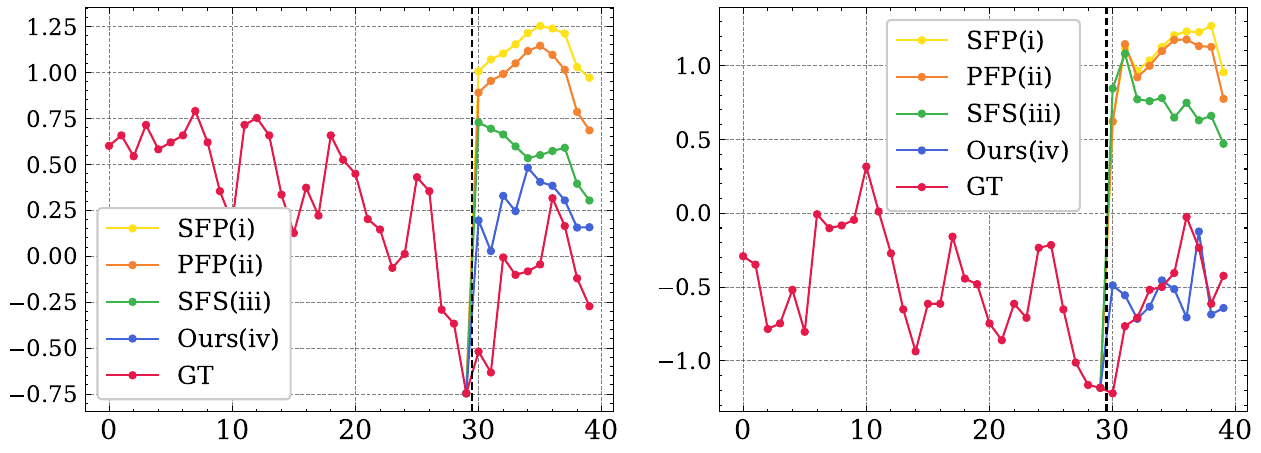}
    % \caption[]{{\small}}    
    % \vspace{-5pt}
    \caption{Display of forecasting results for three paradigms and Seq2Peak \small{(with the red line representing the ground truth)}}
    \label{fig:results}
    % \vspace{-5pt}
\end{figure}

\subsection{Main Results}
We first examine the four paradigms depicted in Figure \ref{fig:4frameworks}, implemented on Transformer. The results of these experiments (Figure \ref{fig:results}) demonstrate the superiority of Seq2Peak over the other three paradigms in accurately predicting future peak-hour series. Given the poor autocorrelation of the peak-hour series mentioned earlier, the methods that directly predict peak-hour (SFP, PFP) fail to capture the trend of the peak-hour series, resulting in underfitting. SFS displayed the best performance among all paradigms, leading us to select it as a strong baseline for subsequent experiments.

Furthermore, we applied our Seq2Peak framework to four commonly used forecasting models\footnote{Our code will be made publicly available at \url{https://github.com/zhangzw16/Seq2Peak}.}. Table \ref{tab:full_results} compares the forecasting accuracy of the baselines and Seq2Peak. The results consistently show that Seq2Peak significantly outperforms all four baselines. Moreover, Seq2Peak demonstrates stable performance, contrasting sharply with the baselines which exhibit a high increase in error with the extension of the prediction length. These experiments affirm that the Seq2Peak framework can effectively enhance the accuracy and robustness of peak-hour series forecasting.

\begin{figure}[thbp]
    \centering  
    \includegraphics[width=0.9\columnwidth]{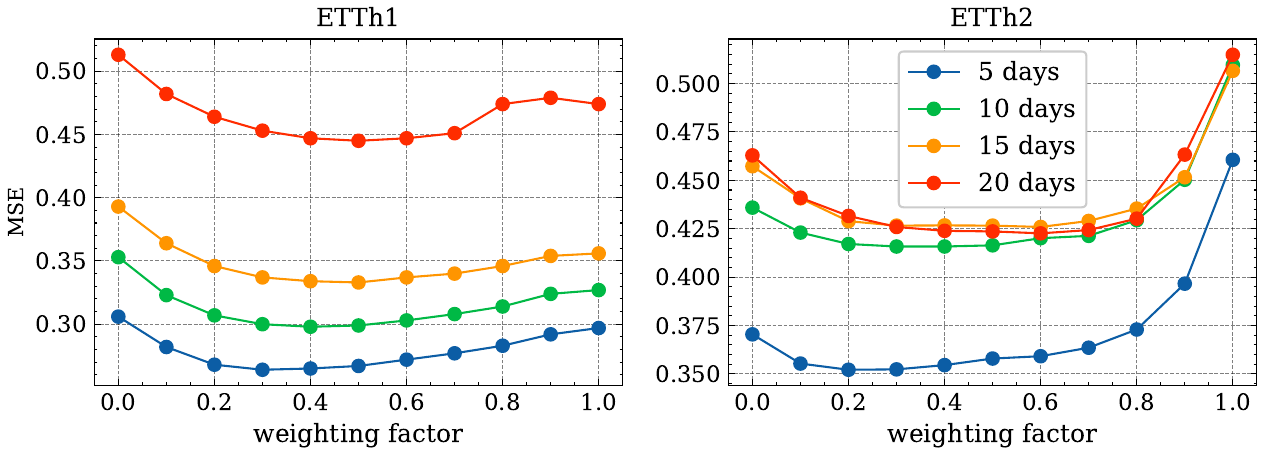}  
    \vspace{-5pt}
    \caption{Effect of peak weighting factor $\alpha$ \small{(tested with DLinear on ETTh1/2)}}
    \label{fig:alpha}
    \vspace{-10pt}
\end{figure}

\input{tables/ablation.tex}
\subsection{Ablation Studies}
We delve deeper into the effectiveness of each module in the proposed framework. Ablation experiments are conducted using the Transformer and DLinear models on the ETT1 dataset. As shown in Figure \ref{tab:ablation_on_architecture}, both CyclicNorm and Seq2Peak Decoder can enhance performance individually, and using them together yields even greater improvements. This demonstrates how these two modules, by focusing on different challenges and empowering the forecasting model from different perspectives, complement each other.

In addition, we provide a hyper-parameter study for $\alpha$ in Eq.\ref{loss func} by plotting dynamic curves. As shown in Figure \ref{fig:alpha}, we examine the performance on ETTh1 and ETTh2 datasets using Seq2Peak enhanced DLinear model. The tendency of the plot indicates that the best-performing $\alpha$ occurs around 0.5, which validates the necessity of applying the hybrid loss function.

\section{Conclusion}
In conclusion, this paper addresses the crucial but often overlooked issue of PHSF. We proposed Seq2Peak, a novel framework tailored for PHSF tasks, bridging the gap between TSF methods and PHSF. Seq2Peak has demonstrated significant performance improvements across datasets for transformer- and non-transformer-based state-of-the-art TSF models. This study effectively solves the PHSF problem and paves the way for further explorations in this critical and challenging field.

\begin{acks}
This research is supported by Huawei Corporation, National Natural Science Foundation of China under Grants U2230201 and 61971266, the Guoqiang Institute of Tsinghua University, and the Clinical Medicine Development Fund of Tsinghua University.
\end{acks}


\bibliographystyle{ACM-Reference-Format}
\balance
\bibliography{references}

% \appendix

\end{document}

%% file: tables/compare.tex
\begin{table*}[t]
	\centering
	\caption{Performance promotion by applying the proposed framework to four TSF models.}\label{tab:full_results}
	% \resizebox{1.0\textwidth}{!}{
    \renewcommand\arraystretch{0.24}
    \begin{threeparttable}
    \begin{small}\begin{tabular}{c|c|cc|cc|cc|cc|cc|cc|cc|cc} 
    \toprule
    \multicolumn{2}{c}{Dataset} & \multicolumn{2}{c}{Transformer} & \multicolumn{2}{c}{\textbf{+Seq2Peak}} & \multicolumn{2}{c}{Informer} & \multicolumn{2}{c}{\textbf{+Seq2Peak}} & \multicolumn{2}{c}{Autoformer} & \multicolumn{2}{c}{\textbf{+Seq2Peak}} & \multicolumn{2}{c}{DLinear} & \multicolumn{2}{c}{\textbf{+Seq2Peak}}\\  
    % \cmidrule(lr){3-4} \cmidrule(lr){5-6} \cmidrule(lr){7-8} \cmidrule(lr){9-10} \cmidrule(lr){11-12} \cmidrule(lr){13-14} \cmidrule(lr){15-16} \cmidrule(lr){17-18}
    \multicolumn{2}{c}{Metric} & MSE & MAE & MSE & MAE & MSE & MAE & MSE & MAE & MSE & MAE & MSE & MAE & MSE & MAE & MSE & MAE\\ 
    \toprule
    \multirow{5}{*}{\rotatebox{90}{\scalebox{0.95}{ETTh1}}}
    &  5 & 2.585 & 1.403 & \boldres{0.299} & \boldres{0.390} & 2.120 & 1.232 & \boldres{0.341} & \boldres{0.410} & 0.486 & 0.519 & \boldres{0.334} & \boldres{0.413} & 0.306 & 0.378 & \boldres{0.240} & \boldres{0.338}\\ 
    & 10 & 2.547 & 1.376 & \boldres{0.330} & \boldres{0.406} & 2.502 & 1.265 & \boldres{0.367} & \boldres{0.420} & 0.588 & 0.549 & \boldres{0.365} & \boldres{0.426} & 0.353 & 0.414 & \boldres{0.263} & \boldres{0.358}\\ 
    & 15 & 2.668 & 1.342 & \boldres{0.353} & \boldres{0.418} & 2.635 & 1.358 & \boldres{0.413} & \boldres{0.451} & 0.617 & 0.561 & \boldres{0.338} & \boldres{0.413} & 0.393 & 0.442 & \boldres{0.279} & \boldres{0.368}\\ 
    & 30 & 2.684 & 1.446 & \boldres{0.439} & \boldres{0.466} & 2.779 & 1.418 & \boldres{0.503} & \boldres{0.507} & 0.747 & 0.633 & \boldres{0.398} & \boldres{0.440} & 0.513 & 0.517 & \boldres{0.335} & \boldres{0.399}\\ 
    % \cmidrule(lr){2-18}
    & \textbf{Avg} & 2.621 & 1.392 & \boldres{0.355} & \boldres{0.420} & 2.509 & 1.318 & \boldres{0.406} & \boldres{0.447} & 0.610 & 0.566 & \boldres{0.359} & \boldres{0.423} & 0.391 & 0.438 & \boldres{0.279} & \boldres{0.366}\\ 
    \midrule

    \multirow{5}{*}{\rotatebox{90}{\scalebox{0.95}{ETTh2}}}
    & 5 & 1.847 & 1.094 & \boldres{0.375} & \boldres{0.448} & 2.223 & 1.278 & \boldres{0.592} & \boldres{0.540} & 0.448 & 0.499 & \boldres{0.400} & \boldres{0.458} & \boldres{0.294} & \boldres{0.382} & 0.301 & 0.384\\ 
    & 10 & 1.932 & 1.111 & \boldres{0.391} & \boldres{0.444} & 1.849 & 1.158 & \boldres{0.540} & \boldres{0.522} & 0.479 & 0.513 & \boldres{0.429} & \boldres{0.471} & 0.376 & 0.428 & \boldres{0.360} & \boldres{0.423}\\ 
    & 15 & 1.873 & 1.103 & \boldres{0.398} & \boldres{0.452} & 1.886 & 1.190 & \boldres{0.557} & \boldres{0.533} & 0.517 & 0.532 & \boldres{0.456} & \boldres{0.487} & 0.426 & 0.459 & \boldres{0.373} & \boldres{0.433}\\ 
    & 30 & 1.946 & 1.186 & \boldres{0.437} & \boldres{0.474} & 2.138 & 1.290 & \boldres{0.646} & \boldres{0.566} & 0.623 & 0.582 & \boldres{0.591} & \boldres{0.547} & 0.664 & 0.582 & \boldres{0.435} & \boldres{0.476}\\ 
    % \cmidrule(lr){2-18}
    & \textbf{Avg} & 1.900 & 1.124 & \boldres{0.400} & \boldres{0.455} & 2.024 & 1.229 & \boldres{0.584} & \boldres{0.540} & 0.517 & 0.532 & \boldres{0.469} & \boldres{0.491} & 0.440 & 0.463 & \boldres{0.367} & \boldres{0.429}\\ 
    \midrule
      
    \multirow{5}{*}{\rotatebox{90}{\scalebox{0.8}{Electricity}}}
    & 5 & 0.457 & 0.469 & \boldres{0.273} & \boldres{0.345} & 0.893 & 0.681 & \boldres{0.307} & \boldres{0.366} & 0.330 & 0.395 & \boldres{0.299} & \boldres{0.361} & 0.252 & 0.326 & \boldres{0.229} & \boldres{0.305}\\ 
    & 10 & 0.551 & 0.513 & \boldres{0.293} & \boldres{0.358} & 1.030 & 0.739 & \boldres{0.327} & \boldres{0.380} & 0.356 & 0.410 & \boldres{0.343} & \boldres{0.388} & 0.286 & 0.351 & \boldres{0.263} & \boldres{0.330}\\ 
    & 15 & 0.524 & 0.510 & \boldres{0.326} & \boldres{0.380} & 1.450 & 0.917 & \boldres{0.335} & \boldres{0.383} & 0.403 & 0.438 & \boldres{0.354} & \boldres{0.399} & 0.312 & 0.371 & \boldres{0.289} & \boldres{0.350}\\ 
    & 30 & 0.557 & 0.517 & \boldres{0.351} & \boldres{0.398} & 1.530 & 0.959 & \boldres{0.379} & \boldres{0.409} & 0.459 & 0.466 & \boldres{0.399} & \boldres{0.429} & 0.373 & 0.415 & \boldres{0.352} & \boldres{0.394}\\ 
    % \cmidrule(lr){2-18}
    & \textbf{Avg} & 0.522 & 0.502 & \boldres{0.311} & \boldres{0.370} & 1.226 & 0.824 & \boldres{0.337} & \boldres{0.385} & 0.387 & 0.427 & \boldres{0.349} & \boldres{0.394} & 0.306 & 0.366 & \boldres{0.283} & \boldres{0.345}\\ 
    \midrule

    \multirow{5}{*}{\rotatebox{90}{\scalebox{0.95}{Traffic}}}
    & 5 & 4.030 & 1.007 & \boldres{2.101} & \boldres{0.814} & 7.148 & 1.729 & \boldres{2.388} & \boldres{0.951} & 3.801 & 1.050 & \boldres{2.253} & \boldres{0.877} & 2.303 & 0.905 & \boldres{2.050} & \boldres{0.836}\\ 
    & 10 & 4.045 & 0.991 & \boldres{2.024} & \boldres{0.796} & 7.458 & 1.796 & \boldres{2.513} & \boldres{0.986} & 3.889 & 1.076 & \boldres{2.254} & \boldres{0.876} & 2.377 & 0.918 & \boldres{2.110} & \boldres{0.844}\\ 
    & 15 & 4.068 & 0.993 & \boldres{2.072} & \boldres{0.808} & 8.138 & 1.933 & \boldres{2.684} & \boldres{1.034} & 3.932 & 1.068 & \boldres{2.458} & \boldres{0.915} & 2.447 & 0.931 & \boldres{2.145} & \boldres{0.848}\\ 
    & 30 & 4.270 & 1.037 & \boldres{2.206} & \boldres{0.834} & 8.881 & 2.071 & \boldres{3.139} & \boldres{1.172} & 3.908 & 1.085 & \boldres{2.513} & \boldres{0.920} & 2.641 & 0.969 & \boldres{2.256} & \boldres{0.868}\\ 
    % \cmidrule(lr){2-18}
    & \textbf{Avg} & 4.103 & 1.007 & \boldres{2.101} & \boldres{0.813} & 7.906 & 1.882 & \boldres{2.681} & \boldres{1.036} & 3.883 & 1.070 & \boldres{2.370} & \boldres{0.897} & 2.442 & 0.931 & \boldres{2.140} & \boldres{0.849}\\ 
    \bottomrule
	\end{tabular}
    \begin{tablenotes}
        \footnotesize
        \item[] We conduct a comprehensive comparison of various competitive models across different prediction lengths, specifically at 5, 10, 15, and 30-day peak-hour series. For all datasets, the input sequence length is consistently set to 30 days. The 'Avg' value is derived from the average of all four prediction lengths.
	\end{tablenotes}
    \end{small}
    
  \end{threeparttable}
  % }
\end{table*}

%% file: tables/ablation.tex
\begin{table}[ht] 

  \caption{Model architecture ablations.}\label{tab:ablation_on_architecture}
  \centering
    \vspace{-5pt}
  \renewcommand\arraystretch{0.3}
    \resizebox{0.9\columnwidth}{!}{
    \small{\begin{threeparttable}
    \begin{tabular}{c|cc|cc|cc}
    \toprule
    Horizons & \multicolumn{2}{c}{5} & \multicolumn{2}{c}{10} & \multicolumn{2}{c}{15} \\
    \cmidrule(lr){2-3} \cmidrule(lr){4-5} \cmidrule(lr){6-7}
    Metrics & MSE & MAE & MSE & MAE & MSE & MAE \\

    \hline
        Transformer
        & 2.285 & 1.403 & 2.547 & 1.376 & 2.668 & 1.342 \Tstrut \\
        + Decoder
        & 2.138 & 1.156 & 1.988 & 1.194 & 0.935 & 0.728 \\ %
        + CyclicNorm
        & \secondres{0.423} & \secondres{0.459} & \secondres{0.421} & \secondres{0.467} & \secondres{0.439} & \secondres{0.480} \\ %
        \textbf{+ Seq2Peak}
        & \boldres{0.299} & \boldres{0.390} & \boldres{0.330} & \boldres{0.406} & \boldres{0.353} & \boldres{0.418} \\ %
        \hline
        DLinear
        & 0.306 & 0.378 & 0.353 & 0.414 & 0.393 & 0.442 \\ %
        + Decoder
        & 0.267 & 0.370 & 0.299 & 0.393 & 0.333 & 0.416 \\ %
        + CyclicNorm
        & \secondres{0.265} & \secondres{0.354} & \secondres{0.294} & \secondres{0.379} & \secondres{0.313} & \secondres{0.392} \\ %
        \textbf{+ Seq2Peak}
        & \boldres{0.240} & \boldres{0.338} & \boldres{0.263} & \boldres{0.358} & \boldres{0.279} & \boldres{0.368} \\ %
        \bottomrule
    \end{tabular}
    \begin{tablenotes}
        \footnotesize
        \item[] The term "\textbf{+ Seq2Peak}" indicates the addition of the complete framework, which includes both the CyclicNorm and Decoder components.
	\end{tablenotes}
  \end{threeparttable}}}
  \vspace{-5pt}
\end{table}